\documentclass[11pt]{article}

\usepackage{stfloats}

\usepackage[preprint]{acl}

\usepackage{times}
\usepackage{latexsym}

\usepackage[T1]{fontenc}
\usepackage[utf8]{inputenc}

\usepackage{microtype}

\usepackage{inconsolata}

\usepackage{graphicx}
\usepackage{amsmath}
\usepackage{amsfonts}
\usepackage{multirow}
\usepackage{fvextra}
\usepackage{arydshln}
\usepackage{comment}

\title{Vroom-Vroom at SHROOM-Visions: A Multi-Judge Committee for Detecting Hallucinated Spans in Vision-Language Outputs}

\author{Toqeer Ehsan, Nico Penttilä, Richard Schmidt, Arash Hajikhani, Victoria Palacin \\
Reliable Intelligence Team, Physical AI, VTT Technical Research Centre of Finland\\ 
\{{\tt firstname.lastname\}}@vtt.fi}

\begin{document}
\maketitle
\begin{abstract}
%This paper describes our submission to the SHROOM-visions shared task on detecting and classifying hallucinated character spans in vision-language model outputs across four languages. We employ several fine-tuned vision-language models and activation probes as independent annotators and combine their span predictions through character-level majority voting. The approach ranks first in three of the four languages and places on the podium in every language and evaluation metric. Our analysis indicates that, in a task constrained by annotator disagreement, aggregating diverse judges outperforms more elaborate weighting and calibration strategies.
This paper describes our submission to the SHROOM-Visions shared task on detecting and classifying hallucinated character spans in vision-language model outputs across four languages. We employ several fine-tuned vision-language models as independent annotators and combine their span predictions through character-level majority voting, and additionally explore activation probes. The approach ranks first in three of four languages and places on the podium in every language and metric. Our analysis indicates that disagreement among diverse models tracks disagreement among human annotators.
\end{abstract}

\section{Introduction}
\label{sec:1}
Vision-Language Models (VLMs) are increasingly used to answer questions, describe scenes, and read text in images. Like text-only models, they often produce fluent and confident outputs that are factually unsupported. In visual settings, such errors take distinct forms: invention, miscounting, relational errors, misinterpretation, or character-recognition errors \cite{liu2024survey,bai2025survey}. Their growing role in deployed systems motivates automatic and span-precise \emph{error detection} as a safeguard \cite{huang2025survey,zhang2025siren}. 

%Identifying these errors at the character level is harder than it appears. Whether a text-span is unsupported often involves subjective judgment, and human annotators frequently disagree both on whether text is hallucinated and on the corresponding span boundaries \cite{mickus2024shroom,vazquez2025mushroom}. This disagreement reflects the inherent ambiguity of the task and places a practical upper bound on detector performance. The SHROOM-visions shared task\footnote{https://helsinki-nlp.github.io/shroom/2026} aims to address this challenge through a multilingual, multi-annotator, span-level benchmark covering English, French, Italian, and Chinese. For each character in a model response, systems must predict its hallucination probability and the associated error type \cite{mickus2026sheep}. Systems are compared for each language using character-level overlap and confidence correlation against the human annotator gold set.

\begin{figure}[ht]
  \centering
  \includegraphics[width=\linewidth]{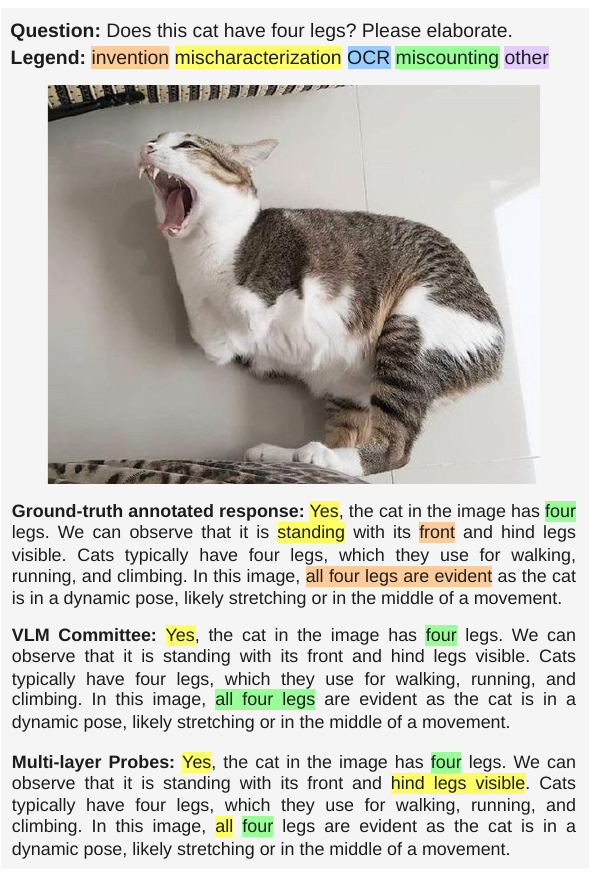}
  \caption{{\small Illustration of hallucination predictions from multi-layer probe and VLM ensemble against human annotations.}}
  \label{fig:1}
\end{figure}

Predicting hallucination at the character level is particularly challenging, as human evaluators likewise disagree on whether a span is hallucinated and where it begins and ends \cite{mickus2024shroom,vazquez2025mushroom}. This disagreement, reflecting inherent ambiguity, sets a practical upper bound on detector performance. We present multiple detection approaches in the context of the SHROOM-Visions \citep{vaquez2026overviewshroomvisions2026shared} shared task\footnote{https://helsinki-nlp.github.io/shroom/2026}, a supervised hallucination detection challenge that is built on a multi-annotator, span-level benchmark covering English, French, Italian, and Chinese \citep{mickus2026sheep}. Systems are compared for each language using character-level overlap and confidence correlation against the human annotator's gold set of real VLM responses. Figure~\ref{fig:1} illustrates a human hallucination annotation alongside model predictions.

We explore two parallel approaches: fine-tuned VLMs treated as independent judges
combined through character-level voting, and VLM activation probes trained on
internal states. Our contributions are twofold: (i) a committee aggregating
multi-model span predictions into character-level probabilities; (ii) evidence
that multi-model diversity reflects annotation disagreement. Our strategy ranks
first in three of four languages. 
%Code and fine-tuned checkpoints are freely available.\footnote{\url{https://github.com/toqeerehsan/vlm_hallucination_detect};  \url{https://huggingface.co/USERNAME}}
%We address the task by combining multiple detectors rather than relying on a single model. We adapt and prompt multiple VLMs and activation probes to identify unsupported spans, then aggregate their predictions through character-level voting. Our contributions are twofold: (i) a committee method that converts model-specific span predictions into calibrated character-level probabilities; (ii) evidence that agreement between two models best reflects the aggregation of gold annotations. The empirical analysis shows that our strategy ranks first in three of four languages and outperforms judge-specific weighting and probability calibration.

\section{Related Work}

\textbf{Vision hallucination benchmarks:} We situate the SHEEP dataset among vision-hallucination datasets along two axes: sourcing of ground truth and hallucination definition. Automated ground truth creation, including synthetic vision-responses \citep{xie2026, zhang-etal-2024-enhancing-hallucination} or error-in-caption augmentation \citep{shekhar-etal-2017-foil}, has been criticized for distributional mismatch \citep{mickus2026sheep}. In contrast, human-sourced annotations of naturally occurring hallucinations, to which SHEEP belongs, can capture authentic and emerging hallucination types, but at the price of annotation and span inconsistencies. This can partially be addressed with the explicitness of the hallucination definition. Existing benchmarks separate descriptive errors, which violate image-grounded features, from errors in visual reasoning \citep{Guan2024, seth-etal-2025-hallucinogen}. With SHEEP belonging to the former descriptive group, additional variation comes with the definition of boundaries, particularly including factually correct elaboration \citep{Gunjal2024}, the handling of hard-to-verify, specialized world knowledge, or the inclusion of decoding errors, such as repetition, template, or code switching errors \citep{Moon2025}, that SHEEP captures under \textit{OTHER}. 

\textbf{Vision hallucination detection:} A smaller body of model-free approaches uses reference captions under a closed-world assumption to identify violations \citep{rohrbach-etal-2018-object} or use between-rollout consistency as uncertainty proxy \citep{zhang-etal-2023-sac3}. In contrast, an extensive body of literature has been concerned with utilizing a secondary VLM to detect hallucinations, either through (i) VLM-as-a-Judge configurations or by (ii) Analysis of Internal Representation. Instruction-following variants include factual decomposition \citep{jing-etal-2024-faithscore}, inter-judge consistency \citep{zhang2024vluncertaintydetectinghallucinationlarge}, or VLM synthesis with smaller attribute models \citep{li-etal-2025-visual}. Interpretability-based methods include uncertainty calibration \citep{geng-etal-2024-survey, li-etal-2024-reference}, which is less effective for hallucinations driven by the language modeling prior \citep{shoby2026overthinkingcauseshallucinationtracing}, but more prominently, supervised methods using visual attention-based attributes \citep{wang2026}, embedding representations \citep{chen2024insidellmsinternalstates}, or between-layer dynamics \citep{Sujoy2026}.

\section{Methodology}

\textbf{EDA and training split:} \label{sec:EDA}
To inform subsequent architecture, training, and prompt design, we performed exploratory data analysis (EDA), surfacing differences at both the hallucination and annotation levels. We find that the annotations are skewed toward invention and mischaracterization as dominant classes (80-88\% of all errors), with severe correlation of error occurrence, where the 10th percentile with the most erroneous responses holds above 50\% of all error characters. Most challenging is inter-annotator disagreement: 80-86\% of hallucination characters are marked by only one of three annotators across languages, motivating the subsequent qualitative analysis of annotation patterns. For the SHEEP dataset, hallucinations are defined as "unsupported by or contradictory to the [...] input image" \citep{mickus2026sheep}, leaving considerable uncertainty regarding both boundary specifications and the verifiability of specialized parametric knowledge. Regarding \textbf{hallucination boundaries}, one inconsistency is the marking of accompanying, factually correct descriptive elaboration of incorrectly identified attributes. Another concerns hard-to-verify but factually correct general knowledge. Prominent examples include the correct inference of geolocations that, while theoretically grounded in the image, are hard to verify without a literal display and were subsequently annotated as hallucinations. Examples are displayed in Figures \ref{fig:hallu_elab} and \ref{fig:hallu_world}. To account for the data distribution and the reuse of source images, we perform a \textit{train, development, test} split that groups by image ID and stratifies by error type on a span, not character level. 

%\subsection{Architectures}
\subsection{Vision-Language Judges}
For the training sets and few-shot examples, we converted each annotation into an inline representation, i.e., <hall label="LABEL" prob="PROB">hallucinated text</hall>. This allowed VLM judges to produce predictions in the same format as their input. We simplified the gold annotations by removing overlapping spans and mapped agreement scores to three probability classes: approximately $0.33$, $0.67$, and $1.0$. We retained samples whose spans are all between one and one hundred characters, excluding unusually long spans. Predicted markup was converted back to character offsets, with each span assigned the probability of its confidence class. For each evaluation item, we retrieved six in-context examples from the training set using weighted cosine similarity.\footnote{Image embeddings: \texttt{openai/clip-vit-large-patch14}, and response and prompt embeddings: \texttt{BAAI/bge-m3}.} We used a 50:30:20 ratio for image, response, and prompt similarity, respectively, and selected demonstrations containing at least three hallucination spans. For fine-tuning, we constructed three distinct training sets to decorrelate the VLM judges.

We used five VLM judges from four backbone families: Gemma-4 \cite{gemma4}, Mistral-Small \cite{mistralsmall31}, Qwen3-VL \cite{qwen3vl}, and a second Qwen variant \cite{qwen36}, providing diverse inductive biases. 
%We obtain the judges through two complementary strategies. 
In the \emph{few-shot} setting, each model received the image, question, and response with six in-context demonstrations retrieved by similarity, and inserted inline markup around unsupported spans. In the \emph{fine-tuning} setting, we adapted each model with LoRA \cite{hu2022lora} on inline-annotated data, using a distinct sample per model. All judges produced responses with identified hallucination labels and confidence probabilities. Prompt templates, retrieval details, and LoRA hyperparameters are provided in Appendix~\ref{app:vlm-configs}.

\subsection{Internal State Probing}
\label{sec:probing}
Probing refers to the family of techniques where language models' internal activations are extracted and analyzed. Prior work has successfully used probes to study representation of values \cite{shen-etal-2025-revisiting} and concepts \cite{abdelwahab-etal-2026-thinking} of LLMs; they are a popular approach in LLM monitoring \cite{NEURIPS2025_b9501301} and existing work has found that internal states are a promising direction for detecting hallucinations \cite{ICLR2025_a712d461, kossen2024semanticentropyprobesrobust, NEURIPS2025_7b8694d5, han2025simple, kim-etal-2025-detecting}. Inspired by these results, our team trained multiple classes of probes of ranging complexity on the internal states of both Qwen3-VL-4B\footnote{https://huggingface.co/Qwen/Qwen3-VL-4B-Instruct} and Qwen3-VL-8B\footnote{https://huggingface.co/Qwen/Qwen3-VL-8B-Instruct}. 

All of the probes feature transformer encoder layers and down projection before the classification heads in order to facilitate horizontal information flow between tokens to better incorporate the span-based nature of the task. We refer to this variant as Multi-layer, since the probe input is a learned mixture of all layers rather than a single selected layer. So given $f^l\in\Re^{n\times D}$ being the hidden state extracted from layer $l$ from the pretrained VLM, the classification heads operate on $z^l=projection(encoder(f^l))$. $z^l\in\Re^{n\times d}, d<<D$ is then fed to classification head(s) to produce per-token hallucination estimations $\hat{y}=head(z_\theta^l),\hat{y}\in\Re^{n\times5}, \hat{y}_{i,j}\in[0,1]$. The probes were optimized with binary cross entropy loss.

We initially trained a linear classification head to estimate per-type probability of the token being hallucinated only on the response token positions. Switching the probe's linear layer to a two-layer perceptron yielded small improvements on the baseline. A large improvement was achieved by compressing the image-token position embeddings to a smaller number of mean-pooled tokens and then cross attending over the mean-pooled representations. Increasing the number of compressed tokens yielded diminishing improvements after 256 tokens. We next added auxiliary heads with their respective losses to the training. Namely, we added start and end boundary logits, sample level logit and a per-token hallucination logit to the existing hallucination type classification head, with the final prediction being gated by the product of the sample level logit and per-token hallucination logit. 
We refer to this variant as Multi-stage, since the type prediction is gated by two preceding decisions rather than produced in a single step. 
Finally, we incorporated the per-position residual stream changes in the input to the probe, inspired by existing work \cite{kim-etal-2025-detecting} and the intuition that hallucinations may manifest as language priors overriding the visually grounded representations. This was implemented by adding a difference of layers $h^l=f^l-f^{l-1}, l\in L$ to the hidden states: $f^L=[[f^0, 0], [f^1, f^1-f^0], ... [f^l, f^l-f^{l-1}]]$, passing $f^L$ through layer-wise encoder $h^L=encoder(f^L)$ from which we acquire scores for each layer $s^l=wh^l+b$ and further pass them through softmax $a^l=\frac{exp(s^l)}{\Sigma_kexp(s^k)}$ which are used to create a weighted average of the layers which gets down-projected and functions as the replacement of $f^l$ when creating the input $z$ to the classification heads: $z^L=projection(encoder(projection(\sum a^lh^l)))$.

\begin{table*}[ht]
\small
\centering
\setlength{\tabcolsep}{2.5pt}
\renewcommand{\arraystretch}{1.1}
\begin{tabular}{ll|cccc|cccc|cccc}
\hline
\multicolumn{2}{l}{\textbf{Approach}} & \multicolumn{4}{c}{\textbf{Cor\_lbl}} & \multicolumn{4}{c}{\textbf{Cor}} & \multicolumn{4}{c}{\textbf{IoU}} \\
\hline
 & \textbf{Model} & \textbf{EN} & \textbf{FR} & \textbf{IT} & \textbf{ZH} & \textbf{EN} & \textbf{FR} & \textbf{IT} & \textbf{ZH} & \textbf{EN} & \textbf{FR} & \textbf{IT} & \textbf{ZH} \\
\hline
 & Gemma-4-ft & 0.3772 & 0.3897 & 0.4153 & 0.4575 & 0.4276 & 0.4556 & 0.4952 & 0.5173 & 0.3734 & 0.4124 & 0.4422 & 0.4751 \\
 & Mistral-small-ft & 0.3384 & 0.3414 & 0.3750 & 0.3912 & 0.3734 & 0.4031 & 0.4625 & 0.4451 & 0.3229 & 0.3589 & 0.4224 & 0.4136 \\
 & Qwen3.6-ft & 0.3738 & 0.3759 & 0.3729 & 0.4359 & 0.4127 & 0.4287 & 0.4445 & 0.4774 & 0.3691 & 0.3842 & 0.3935 & 0.4395 \\
 & Qwen3-vl-ft-3shot & \textbf{0.3904} & 0.3639 & 0.3751 & 0.4497 & 0.4344 & 0.4272 & 0.4729 & 0.5042 & 0.3721 & 0.3825 & 0.4166 & 0.4599 \\
 \multirow{-5}{*}{\rotatebox{90}{VLMs}} & Gemma-4-6shot & 0.3117 & 0.3329 & 0.3222 & 0.4155 & 0.3890 & 0.4149 & 0.4256 & 0.4991 & 0.3706 & 0.4099 & 0.4207 & 0.4659 \\
\hdashline
& Votes $\geq$ 1 & 0.3227 & 0.3569 & 0.3671 & 0.3927 & 0.4125 & 0.4722 & 0.4911 & 0.4944 & 0.3575 & 0.4230 & 0.4225 & 0.4421 \\
 & Votes $\geq$ 2 & 0.3872 & 0.3979 & \textbf{0.4246} & 0.4669 & \textbf{0.4594} &\textbf{0.4827} & \textbf{0.5355} & \textbf{0.5407} & \textbf{0.3932} & \textbf{ 0.4322} & \textbf{0.4756} & \textbf{0.4936} \\
 & Votes $\geq$ 3 & {0.3903} & \textbf{0.4019} & {0.4125} & \textbf{0.4709} & 0.4437 & 0.4776 & 0.5047 & 0.5229 & 0.3786 & 0.4145 & 0.4443 & 0.4737 \\
\multirow{-4}{*}{\rotatebox{90}{Committee}} & Votes $\geq$ 4 & 0.3803 & 0.3755 & 0.3999 & 0.4575 & 0.4220 & 0.4221 & 0.4665 & 0.4938 & 0.3588 & 0.3664 & 0.4075 & 0.4496 \\
\hline
 & Multi-layer (4B) & 0.3176 & 0.3230 & 0.3410 & 0.3903 & 0.4044 & 0.4034 & 0.4503 & 0.4864 & 0.3801 & 0.3835 & 0.4017 & 0.4551 \\
 & Multi-stage (4B) & 0.3314 & 0.3267 & 0.3429 & 0.4007 & 0.4101 & 0.4087 & 0.4538 & 0.4881 & 0.3813 & 0.3929 & 0.4033 & 0.4607 \\
 & Multi-layer (8B) & 0.3170 & 0.2967 & 0.3251 & 0.3940 & 0.3940 & 0.4133 & 0.4558 & 0.4842 & 0.3744 & 0.3978 & 0.4081 & 0.4594 \\
\multirow{-4}{*}{\rotatebox{90}{Probes}} & Multi-stage (8B) & 0.3236 & 0.3242 & 0.3404 & 0.4076 & 0.3933 & 0.4290 & 0.4655 & 0.4977 & 0.3764 & 0.4091 & 0.4173 & 0.4616 \\
 \hline
\end{tabular}
\begin{comment}
\caption{{\small Language-wise scores against all three metrics on the test split of SHROOM-visions' train set.}}
\end{comment}
\caption{\small Language-wise scores against all three metrics on the test split of SHROOM-Visions' train set. Multi-layer probes read a learned softmax-weighted combination of all layers including residual-stream differences; Multi-stage probes add boundary, sample-level, and token-level heads and gate the type prediction on the latter two (§\ref{sec:probing}).}
\label{tab:dev-results}
\end{table*}

\subsection{Committee Aggregation}
Judges often disagree on span boundaries, so we first project each judge's spans
onto characters: for character $c$, judge $j$'s confidence $p_j(c)$ and label
$\ell_j(c)$ are those of its highest-confidence covering span, and $p_j(c)=0$
where no span covers $c$. Judges can then be compared position by position
irrespective of their segmentation. We count the judges flagging each character,
$v(c)=\sum_{j=1}^{N}\mathbb{I}_{j}(c)$, and sum their confidences:

\vspace{-1.1em}
\begin{equation}
\small
\mathrm{prob}[c] = \frac{1}{N}\sum_{j=1}^{N}\mathbb{I}_{j}(c)p_j(c),
\label{eq:aggregation}
\end{equation}
\vspace{-0.9em}

where $\mathbb{I}_{j}(c)$ indicates whether judge $j$ flags $c$ and $N$ is the
number of judges. Dividing by $N$ combines agreement with confidence;
normalizing by $v(c)$ would give the same value to a character flagged by one
judge and by five, and the finer probability scale benefits the rank-based
metrics. The predicted label is the category with the highest summed confidence.
We retained characters with $v(c)\ge 2$ and merged maximal runs sharing a label
and probability into spans, mirroring the gold labels, the union of three
annotators. The rule is deliberately inclusive, since most gold spans carry only
one annotator. Adding high-recall probes did not improve performance.

\section{Results and Discussion}
We evaluate using three metrics: correlation between predicted and empirical per-character hallucination probabilities (Cor), its per-label variant (Cor\_lbl), and character-level intersection-over-union (IoU). Table~\ref{tab:dev-results} reports per-language results on the labeled test split for individual judges, committee variants with different vote thresholds and the best performing probes. No judge performs best across all settings. Fine-tuned Gemma-4 and Qwen3-VL are strongest in most languages, but different judges lead on specific metrics. Probe scores are closely correlated with one another, suggesting that performance depends on the underlying activations rather than the probe architecture. 

The committee outperforms every individual judge on Cor and IoU, the only
exception being English Cor\_lbl, where Qwen3-vl-ft-3shot scores slightly higher. With \mbox{votes $\geq$ 2}, it achieves the best Cor and IoU across all four languages, showing that aggregation recovers recall lost by fine-tuned judges while suppressing single-judge false positives. The threshold shows a consistent pattern: the two-vote setup performs best for Cor and IoU, whereas the stricter \mbox{votes $\geq$ 3} slightly improves Cor\_lbl in three of four languages by increasing label precision at the cost of coverage. The addition of the probes to the ensemble does not yield improvements overall due to the high correlation of errors with existing approaches.

\subsection{Evaluation on the Challenge Set}
Table~\ref{tab:test-results} shows the scores of the VLM committee on the unseen challenge set from the shared task for all languages. The \mbox{votes $\geq$ 2} configuration ranks first in French, Italian, and Chinese on the correlation metrics and second in English, placing on the podium for every language and metric. English performs the worst, as its longer responses reduce annotator agreement and limit achievable scores. Although the margin on the seen split is small, the \mbox{votes $\geq$ 3} configuration degrades performance for every language and metric on the challenge set, indicating a genuine difference rather than noise.

\begin{table}[ht]
\small
\centering
\setlength{\tabcolsep}{3pt}
\renewcommand{\arraystretch}{1.0}
\begin{tabular}{cl|cccc|c}
%\hline
%\textbf{Cat.} & & \multicolumn{4}{c}{\textbf{Cor\_lbl}} & \multicolumn{4}{c}{\textbf{Cor}} & \multicolumn{4}{c}{\textbf{IoU}} \\
\hline
%\hline
\textbf{Votes} &\textbf{Metrics} & \textbf{EN} & \textbf{FR} & \textbf{IT} & \textbf{ZH} & \textbf{Avg.}\\
\hline
&Cor\_lbl & 0.4251 & 0.4738 & 0.4511 & 0.5040 & 0.4635\\
&Cor & 0.5334 & 0.5873 & 0.5628 & 0.6083 & 0.5730 \\
\multirow{-3}{*}{\rotatebox{90}{$\geq$ 2}} & IoU & 0.4486 & 0.5150 & 0.4793 & 0.5344 & 0.4943\\
\hline
&Cor\_lbl & 0.4064 & 0.4509 & 0.4393 & 0.4988 & 0.4489 \\
&Cor & 0.4965 & 0.5417 & 0.5300 & 0.5812 & 0.5373\\
\multirow{-3}{*}{\rotatebox{90}{$\geq$ 3}}&IoU & 0.4063 & 0.4644 & 0.4447 & 0.5112 & 0.4566 \\
\hline
\end{tabular}
\caption{{\small Language-wise scores on the SHROOM-Visions challenge set from the VLM judge committee with votes $\geq$ 2 and $\geq$ 3. Scores with votes $\geq$ 2 are from shared task rankings.}}
\label{tab:test-results}
\end{table}

\subsection{Label-wise Analysis}

\begin{table}[ht]
\footnotesize
\centering
\setlength{\tabcolsep}{2.7pt}
\renewcommand{\arraystretch}{1.0}
\begin{tabular}{ll|ccccc}
\hline
& \textbf{System} & \textbf{Inv.} & \textbf{MisCh.} & \textbf{OCR} & \textbf{MisC.} & \textbf{Oth.} \\
\hline
& Gemma-4-ft       & 0.208 & 0.178 & 0.446 & \textbf{0.409} & 0.082  \\
& Mistral-small-ft & 0.144 & 0.157 & 0.249 & 0.266 & \textbf{0.106} \\
& Qwen3.6-ft       & 0.182 & 0.112 & 0.399 & 0.365 & 0.036  \\
& Qwen3-vl-ft      & 0.186 & 0.176 & 0.280 & 0.365 & 0.039 \\
& Gemma-4-6shot    & 0.194 & 0.181 & 0.379 & 0.320 & 0.057 \\
\multirow{-6}{*}{\rotatebox{90}{Cor\_lbl}}
& Committee ($\geq$2)    & \textbf{0.235} & \textbf{0.208} & \textbf{0.449} & 0.399 & 0.089  \\
\hline
& Gemma-4-ft       & 0.335 & 0.316 & 0.502 & 0.409 & 0.204  \\
& Mistral-small-ft & 0.282 & 0.259 & 0.341 & 0.266 & 0.231  \\
& Qwen3.6-ft       & 0.276 & 0.231 & 0.534 & 0.364 & 0.133  \\
& Qwen3-vl-ft      & 0.324 & 0.293 & 0.430 & 0.362 & 0.208  \\
& Gemma-4-6shot     & 0.338 & 0.329 & 0.454 & 0.383 & 0.273  \\
\multirow{-6}{*}{\rotatebox{90}{Cor}}
& Committee ($\geq$2) & \textbf{0.397} & \textbf{0.363} & \textbf{0.550} & \textbf{0.441} & \textbf{0.296} \\
\hline
& Gemma-4-ft       & 0.176 & 0.144 & \textbf{0.406} & \textbf{0.342} & 0.095  \\
& Mistral-small-ft & 0.116 & 0.131 & 0.228 & 0.212 & \textbf{0.109}  \\
& Qwen3.6-ft       & 0.143 & 0.091 & 0.369 & 0.309 & 0.036  \\
& Qwen3-vl-ft      & 0.148 & 0.145 & 0.244 & 0.300 & 0.036  \\
& Gemma-4-6shot    & \textbf{0.203} & 0.145 & 0.338 & 0.246 & 0.063 \\
\multirow{-6}{*}{\rotatebox{90}{IoU}}
& Committee ($\geq$2)   & 0.199 & \textbf{0.168} & 0.404 & 0.315 & 0.086 \\
\hline
& Gemma-4-ft       & 0.332 & 0.342 & 0.379 & 0.334 & 0.331 \\
& Mistral-small-ft & 0.361 & 0.357 & 0.440 & 0.446 & \textbf{0.328} \\
& Qwen3.6-ft       & 0.352 & 0.366 & 0.378 & 0.349 & 0.355 \\
& Qwen3-vl-ft      & 0.348 & 0.342 & 0.422 & 0.356 & 0.354 \\
& Gemma-4-6shot    & 0.498 & 0.466 & 0.447 & 0.475 & 0.394 \\
\multirow{-6}{*}{\rotatebox{90}{MAE$_p$$\downarrow$}}
& Committee ($\geq$2)  & \textbf{0.309} & \textbf{0.318} & \textbf{0.310} & \textbf{0.305} & 0.336 \\
\hline
\end{tabular}
\caption{{\small Label-wise scores on the test split of SHROOM-Visions' train set. Cor\_lbl, Cor and IoU are higher-is-better; MAE$_p$ is the mean absolute difference between predicted and gold character probabilities (lower is better). %\textbf{Mis.}~=~miscounting,
%\textbf{Oth.}~=~other. Best per column in bold.}
}}
\label{tab:label-wise}
\end{table}

Table~\ref{tab:label-wise} breaks the official metrics down by hallucination label and adds MAE$_p$, the mean absolute difference between predicted and gold character probabilities. The committee is strongest overall on all four measures, most clearly on the correlation metrics: it obtains the best Cor on every label and the best Cor\_lbl on three of five, raising Cor. A single judge emits only three confidence values, confining it to a coarse grid, whereas averaging over the committee yields a finer range closer to the empirical annotator agreement, reducing the MAE$_p$. The picture is mixed for IoU, which depends only on which characters are marked: individual judges remain competitive on invention, OCR and miscounting, and the committee's advantage comes from suppressing spans that only one judge proposes rather than from marking more text. Performance on \emph{other} is low throughout, consistent with it being the least frequent and least consistently annotated category.

\subsection{VLM Judge vs.\ Annotator Agreement}
Judge agreement tracks human agreement. Table~\ref{tab:votes-annot} groups
characters by how many judges flagged them. The mean number of annotators who
marked them rises monotonically from $0.11$ at zero votes to $2.07$ at five,
with a Spearman correlation of $0.39$ over characters and $0.51$ at the record
level. This supports our claim that model diversity reflects annotator
disagreement and motivates the two-vote ($\geq$2) threshold.

\begin{table}[ht]
\small
\centering
\setlength{\tabcolsep}{5pt}
\renewcommand{\arraystretch}{1.0}
\begin{tabular}{c|cc}
\hline
\textbf{Judges flagging} & \textbf{Mean annotators} & \textbf{Characters} \\
\hline
0 & 0.109 & 633,649 \\
1 & 0.453 & 86,999 \\
2 & 0.743 & 19,838 \\
3 & 1.205 & 9,171 \\
4 & 1.535 & 5,968 \\
5 & 2.072 & 5,260 \\
\hline
\end{tabular}
\caption{{\small Mean number of annotators marking a character, grouped by how
many judges flagged it.}}
\label{tab:votes-annot}
\end{table}

\subsection{Committee vs. Probes}
The two approaches trade accuracy against cost: the committee needs five forward
passes through large fine-tuned VLMs, while a probe adds only a lightweight head
over hidden states from a single pass. Probes match individual judges on IoU but
trail the committee on the correlation metrics, making them preferable when
the inference budget is the binding constraint.

\section{Conclusion}
We present a committee approach to multilingual hallucination span detection in vision-language models. Fine-tuned and few-shot judges independently identify unsupported spans, which are combined through character-level voting. Only corroborated spans are retained, while averaged confidences provide the probabilities used by the correlation metric. The system ranks first in three of four languages.
The results show that activation probes can detect hallucinations from language model hidden states, but the results are highly dependent on the model from which the activations are collected. Currently, it is unclear whether probes offer mechanistic insight into hallucination or merely exploit surface linguistic features.

\section*{Limitations}
Our committee inherits the limitations of its individual judges. Long, fluent hallucinations not confidently detected by multiple models may fall below the two-vote threshold and remain undetected, particularly in the longer English responses where performance is weakest. We also do not explore image preprocessing: images vary considerably in resolution and are passed to the judges without normalization or aspect-ratio-aware transformation, which may affect grounding. The voting threshold is tuned on the labeled test split and applied unchanged to the challenge set, but may not generalize to other models, languages, or annotation schemes. Confidence scores are based on three discrete agreement classes rather than continuous estimates, limiting how precisely uncertainty can be represented. Finally, our analysis is limited to the four task languages and selected judges, and the aggregated predictions may not capture the full complexity of human annotator disagreement.

\section*{Acknowledgments}
This research was part of the Postdoctoral Programme for Research Institutes in Finland, funded by the Finnish Government.
We used Claude (Anthropic) to assist with drafting and editing the paper and with code generation. All decisions, experiments, and results were designed and verified by the authors.

\section*{Data and Code Availability}
Our VLM committee code is available at \url{https://github.com/toqeerehsan/vlm_hallucination_detect} and the fine-tuned checkpoints(v3) at \url{https://huggingface.co/QSTS-VTT}. The SHEEP dataset is distributed by the task organizers.

\bibliography{custom}

\appendix
\section{SHEEP examples: penalizing elaboration and world knowledge} \label{app:sample_hallu_elabrorate}

\begin{figure}[h]
\centering
 \noindent\includegraphics[height=5.5cm]{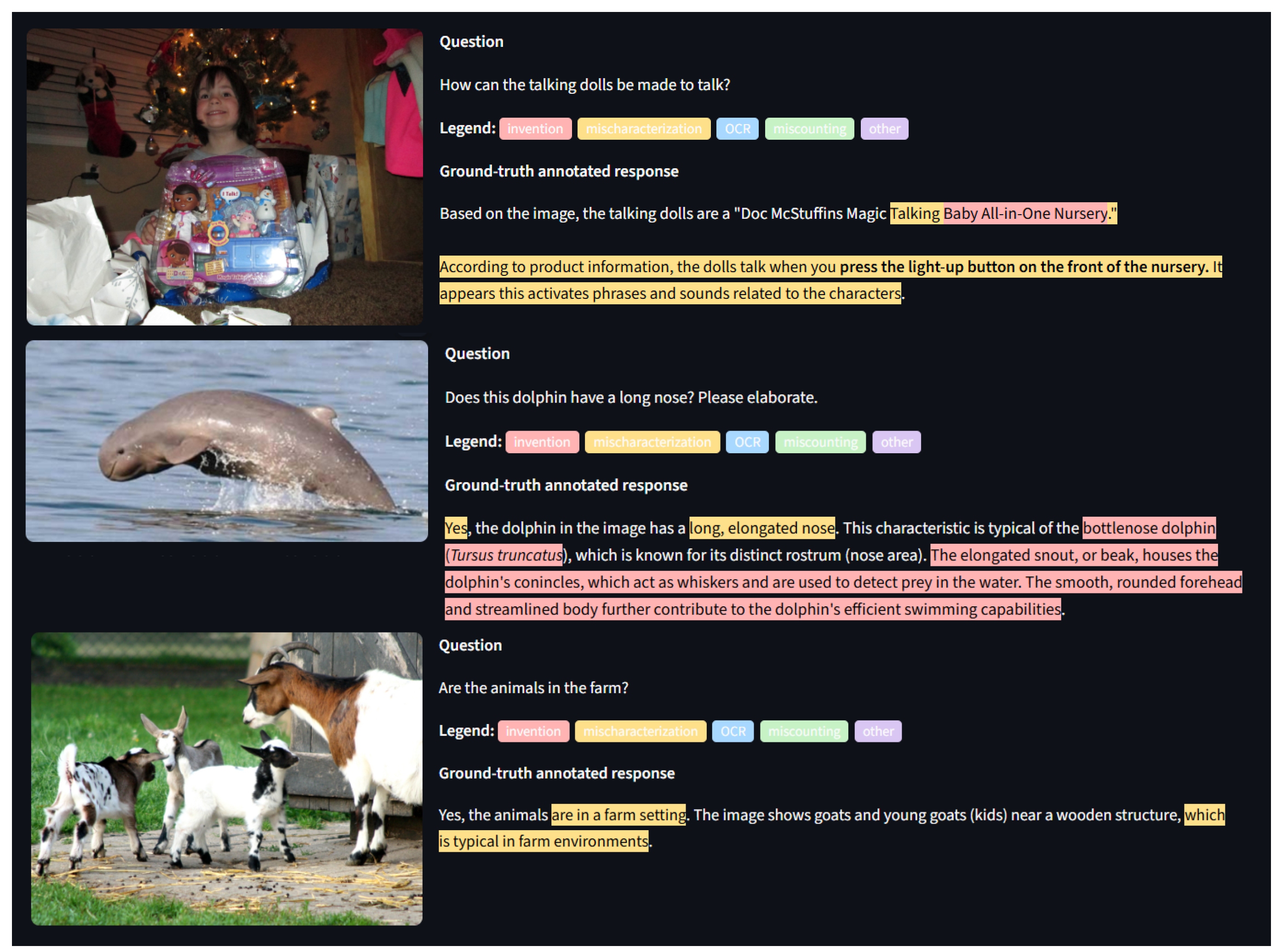} 
 \caption{{\small The three selected samples from the SHEEP dataset, showing penalization of factually correct elaboration for hallucinated attributes.}}
 \label{fig:hallu_elab}
\end{figure}

See Figure~\ref{fig:hallu_elab}.

\section{SHEEP examples: penalizing world knowledge} \label{app:sample_hallu_world_knowledge}

\begin{figure}[h]
\centering
 \noindent\includegraphics[height=5.5cm]{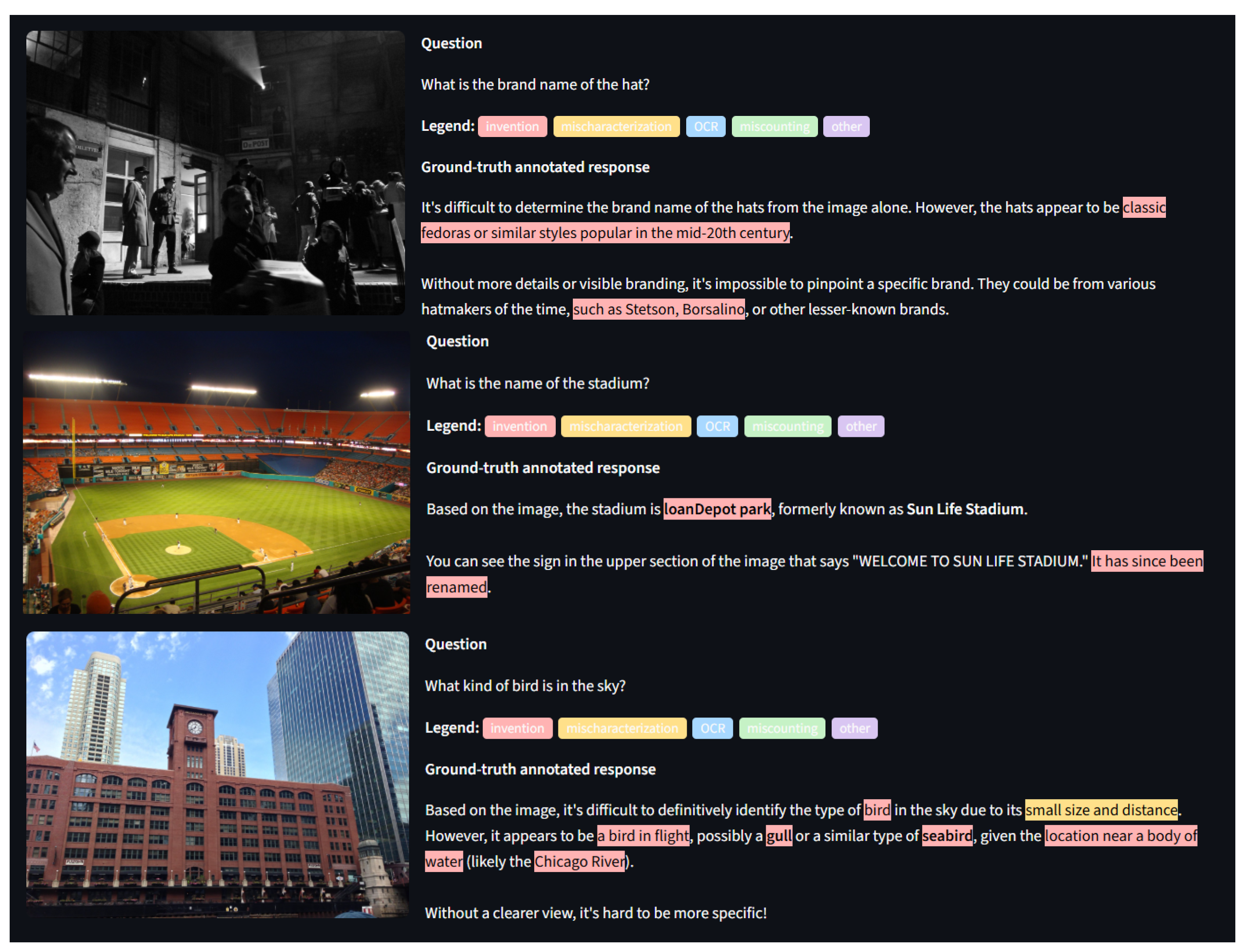} 
 \caption{{\small The three selected samples from the SHEEP dataset, showing penalization of factually correct, but hard to verify world knowledge.}}
 \label{fig:hallu_world}
\end{figure}

See Figure~\ref{fig:hallu_world}.

\section{Probing details}
\label{probing_details}

\subsection{Base model initialization}
The underlying Qwen models from which the embeddings were extracted from were initialized with the image from the dataset, a system prompt, a user prompt consisting of the prompt from the dataset and the response from the dataset.

The system prompt used for extracting the hidden states was the following.
\begin{Verbatim}[fontsize=\footnotesize, breaklines=true, breakanywhere=true, breaksymbolleft={}, breaksymbolright={}]
You are an image-grounded hallucination detector.

Given an image, a query and a response, identify hallucinated spans in the response and assign labels.

Labels:
A. invention : entities, objects, properties, or events not present in the image.
B. mischaracterization : incorrect description of content that is visible.
C. OCR : misreading of text visible in the image.
D. miscounting : incorrect reporting of quantities of visible items.
E. other : the hallucination does not fit in classes A-D.
\end{Verbatim}

The user prompt contained the inputs, with most importantly the response being started and ended with easily identifiable token sequences.

\subsection{Training}

The probes were trained with a learning rate of $1\times10^{-4}$, with a batch size of 24 and with weight decay of 0.01. The weight factors for the different BCE-losses were 1.0 for hallucination type loss, 0.1 for the boundary (start/end) loss, 0.5 for the sample level hallucination loss and 0.25 for the token-level hallucination loss.

\section{VLM judge configurations}
\label{app:vlm-configs}

%\subsection{Retrieval Details}
%\label{app:retrieval}
%For each evaluation item, we retrieve six same-language demonstrations from the training set. Candidates are ranked by a weighted sum of image ($0.50$), response ($0.30$), and prompt ($0.20$) cosine similarities, with each channel min--max normalized per item. Images are embedded with \texttt{openai/clip-vit-large-patch14} and text with \texttt{BAAI/bge-m3}. From the ten highest-ranked candidates, we select six containing at least three hallucination spans. Only spans of $1$--$100$ characters are retained. We construct three distinct subsets so that the fine-tuned judges receive different training samples.

\subsection{Fine-tuning}
\label{app:config}

We adapt each backbone with LoRA \cite{hu2022lora}, using rank $64$, $\alpha=16$, and dropout $0.05$, together with rsLoRA and LoRA+ at a ratio of $4$. Training runs for two epochs with a learning rate of $5\!\times\!10^{-5}$, warmup ratio $0.05$, batch size $1$, gradient accumulation over $8$ steps, and a maximum sequence length of $6192$. We use bf16 without quantization. The judges are based on Gemma~4, Mistral-Small, Qwen3-VL \cite{qwenvl2025}, and Qwen3.6 \cite{qwen36}, with each fine-tuned on a distinct data subset. The committee also includes one few-shot Gemma~4 judge. %\emph{[Insert the exact Gemma model size and number of demonstrations.]}

\subsection{Prompt}
%\begin{figure}[t]
\begin{Verbatim}[fontsize=\footnotesize, breaklines=true, breakanywhere=true, breaksymbolleft={}, breaksymbolright={}]
You are a hallucination-span annotator for vision-language model outputs.

You are given: an IMAGE, a PROMPT that was asked about it, and a RESPONSE that a vision-language model produced. Parts of the RESPONSE are often hallucinated — that is, not supported by the image.

Your job: MARK hallucinations with <hall> tags inserted around every hallucinated span and return the RESPONSE text EXACTLY as given, character for character. Change NOTHING else — do not fix typos, spacing, markdown, or line breaks. Do not add words. Do not add any preamble, reasoning, or explanation. Output ONLY the annotated response text. 

Tag format:
  <hall label="LABEL" prob="PROBCLASS">hallucinated text</hall>

LABEL is exactly one of (never "None", never empty): 
invention - entity/object/property/event not present in the image
mischaracterization - content is visible but described incorrectly
OCR - text visible in the image is misread
miscounting - a quantity of visible items is reported incorrectly
other - a hallucination that fits none of the above

PROBCLASS is exactly one of prob1..prob3 (never "None"): how likely a panel of three human annotators would agree the span is hallucinated (higher = clearer):
prob1 - ~0.3333 (weak) , prob2 - ~0.6667 , prob3 - ~1.0 (blatant).

How to annotate:
- Be thorough: check every claim against the image (colors, counts, object/species identity, brands, text, spatial relations, materials, details) and MARK anything not clearly supported.
- Mark a doubtful span with low-confidence (prob1) class rather than leaving unmarked.
- Mark the SMALLEST text carrying the error (a word or short phrase), not the whole sentence.
- Tags may nest (in rare cases), but never partially cross, never repeat the same tag on the same text, and never wrap markdown like **.
- Leave the response untagged ONLY if every claim is supported.
- Keep the response in its original language.
\end{Verbatim}
%\caption{System prompt used by the judges.}
%\label{fig:prompt}
%\end{figure}

% \begin{figure}[t]
% \small\ttfamily
% [paste the exact SYSTEM_WITH_LABELS prompt here]
% \caption{System prompt used by the judges.}
% \label{fig:prompt}
% \end{figure}

\end{document}